%% file: main.tex
\documentclass[sigconf,natbib=true]{acmart}

\AtBeginDocument{%
  }

\usepackage{booktabs}
\usepackage{tabularx}
\usepackage{siunitx}
\copyrightyear{2026}
\acmYear{2026}
\setcopyright{cc}
\setcctype{by}
\acmConference[CIKM '26]{Proceedings of the 35th ACM International Conference on Information and Knowledge Management}{November 07--11, 2026}{Rome, Italy}
\acmBooktitle{Proceedings of the 35th ACM International Conference on Information and Knowledge Management (CIKM '26), November 07--11, 2026, Rome, Italy}
\acmDOI{10.1145/3799682.3840066}
\acmISBN{979-8-4007-2539-5/2026/11}

\begin{document}

\title[Performance and Stability of LLM Rerankers in CRS]{Retrieval, Scoring, and Decoding Shape Performance and Stability in LLM-based Conversational Recommendation}

\author{Ante Kapetanovic}
\authornotemark[1]
\email{ankapetanovic@infobip.com}
\affiliation{%
  \institution{Infobip}
  \city{Split}
  \country{Croatia}
}

\author{Tomislav Duricic}
\authornotemark[1]
\email{tduricic@infobip.com}
\affiliation{%
  \institution{Infobip}
  \city{Zagreb}
  \country{Croatia}
}

\author{Andro Mercep}
\email{amercep@infobip.com}
\affiliation{%
  \institution{Infobip}
  \city{Zagreb}
  \country{Croatia}
}

\author{Emanuel Lacic}
\email{emlacic@infobip.com}
\affiliation{%
  \institution{Infobip}
  \city{Zagreb}
  \country{Croatia}
}

\renewcommand{\shortauthors}{Ante Kapetanovic, Tomislav Duricic, Andro Mercep, and Emanuel Lacic}


\begin{abstract}
  Large language models (LLMs) are increasingly used as rerankers in conversational recommender systems, yet measured gains depend strongly on the retrieval and inference protocol.
  On the ReDial conversational movie recommendation benchmark, we compare proprietary, open-weight, and fine-tuned LLM rerankers with collaborative-filtering and sequential baselines in a shared retrieve-then-rerank pipeline.
  We vary candidate-pool size, first-stage retriever, and decoding temperature.
  With a shared semantic top-250 candidate pool and strict candidate-aware scoring, the best proprietary reranker reaches NDCG@10 of 0.1497, compared with 0.0939 for the strongest non-LLM baseline.
  The same reranker reaches 0.2925 in zero-shot generation, showing that unconstrained scoring can yield a much larger apparent advantage than matched-pool evaluation.
  No evaluated open-weight LLM outperforms the tuned shallow autoencoder baseline under this protocol.
  For the strongest proprietary and open-weight rerankers, switching from semantic to collaborative-filtering candidates raises NDCG@10 by more than 50\%, showing that measured reranker performance is highly sensitive to candidate generation.
  For the best proprietary reranker, raising temperature from 0 to 1.0 increases top-10 Jaccard distance from 0.0900 to 0.1240 while mean NDCG@10 changes negligibly, whereas weaker LLMs show larger degradation.
  These ReDial results support treating candidate generation, candidate-pool size, scoring policy, and decoding configuration as required reporting fields rather than implementation details.
\end{abstract}


\maketitle

\input{sections/01_introduction}
\input{sections/02_methodology}
\input{sections/03_results}
\input{sections/04_discussion}

\begin{acks}
This research was supported in part by the project Infobip Global Communication Platform (PK.1.1.07.0001), part of the Important Project of Common European Interest on Next Generation Cloud Infrastructure and Services (IPCEI-CIS) consortium.
\end{acks}

\section*{GenAI Usage Disclosure}
Generative AI tools assisted in a supporting capacity: Claude Code (Opus 4.7) for code implementation and data analysis, ChatGPT 5.5 for \LaTeX{} editing and grammar checking, and Google Scholar Labs for related-work identification.
All AI-assisted outputs were reviewed and verified by the authors, who take full responsibility for the work.

\bibliographystyle{ACM-Reference-Format}
\bibliography{references}

\end{document}

%% file: sections/01_introduction.tex
\section{Introduction}
\label{sec:introduction}

Conversational recommender systems (CRS) infer preferences from dialogue rather than from dense interaction histories~\cite{jannach2021survey}.
Large language models (LLMs) have shown promising zero-shot ranking ability on recommendation tasks~\cite{hou2024llmrank} and can also be fine-tuned for the role~\cite{bao2023tallrec}, which is why recent CRS work increasingly relies on them at the ranking stage.
Reported gains over earlier baselines are often substantial, but those numbers depend on pipeline choices that many tend to under-report, i.e., which items are retrieved as candidates, how many of them are shown to the LLM, whether evaluation restricts credit to the candidate set or scores any generated catalog title, and how the model is decoded.
On a single benchmark, these choices can move ranking quality by more than the gap separating top systems, so the same model can appear strong or mediocre depending on how it is evaluated.

Two-stage retrieve-then-rerank is standard for LLM-based ranking~\cite{sun2023rankgpt}, and recent LLM rerankers add user-preference retrieval or graph signals on top of an explicit candidate stage~\cite{zhang2025ur4rec,wei2024llmrec}.
Yet, LLM CRS comparisons are seldom controlled for candidate generation, and reproducibility studies show that recommender comparisons are sensitive to such protocol choices as well as to baseline strength~\cite{dacrema2019progress,krichene2020sampled}.
Stochastic decoding further makes LLM outputs variable across repeated samples and sensitive to the order in which candidates appear in the prompt~\cite{ma2025stable,bito2026invarirank}, yet list-level stability is rarely reported alongside accuracy in LLM CRS evaluations.

We study how these factors jointly shape observed effectiveness on ReDial~\cite{li2018towards}, a CRS benchmark of seeker-recommender movie dialogues in which preferences must be inferred from the conversation alone.
We compare proprietary, open-weight, and fine-tuned LLM rerankers against collaborative-filtering (CF) and sequential rerankers in a shared two-stage pipeline{\renewcommand{\thefootnote}{\fnsymbol{footnote}}\footnotetext[1]{Equal contribution.}}\footnote{Code, prompts, configurations, and outputs: \url{https://github.com/infobip/crs-performance}}.
Matched-pool comparisons hold the retrieved candidates fixed while we vary candidate-pool size, first-stage retriever, and decoding temperature.
We answer the following research questions:

\begin{itemize}
  \item[\textbf{RQ1}] Under a matched semantic candidate pool, how do LLM rerankers compare to CF and sequential rerankers?
  \item[\textbf{RQ2}] How sensitive is LLM reranking quality to candidate-pool size, from zero-shot generation to full-catalog reranking?
  \item[\textbf{RQ3}] How does the choice of first-stage retriever (i.e., semantic, CF, or sequential) affect the quality of an LLM reranker?
  \item[\textbf{RQ4}] How stable are LLM recommendation lists under decoding-temperature sampling, in terms of both ranking quality and list-level agreement?
\end{itemize}

Four proprietary LLMs significantly outperform EASE (0.0939), led by Claude Opus 4.6 at 0.1497, while every evaluated open-weight reranker falls below it.
Expanding the semantic pool from 250 items to the full catalog raises proprietary NDCG@10 by 57--89\%.
Replacing semantic with EASE candidates raises NDCG@10 by 52--59\% across the two tested rerankers.
With higher temperature, Claude Opus 4.6 maintains mean accuracy while Jaccard distance@10 rises from 0.0900 to 0.1240, whereas for Llama-3.3-70B it rises from 0.0230 to 0.7600.
These results identify retrieval strategy, candidate-pool size, scoring policy, and decoding configuration as core experimental variables.

%% file: sections/02_methodology.tex
\section{Methodology}
\label{sec:methodology}

\subsection{Dataset and Task}

We evaluate on ReDial's standard test split (1,025 dialogues) and 6,924-movie catalog~\cite{li2018towards}.
For each dialogue, accepted recommendations are masked and used as ground-truth targets.
Each model returns ten ranked movie titles from family-specific inputs.

LLM prompts contain liked titles, the masked dialogue, and, outside zero-shot generation, a shuffled candidate list.
We label settings by candidate count: c\(K\) denotes \(K\) candidates, c0 no candidate list, and cAll the full catalog.
Candidate-aware prompts require outputs to use only listed titles.
For CF models, liked movies form the implicit interaction history, while accepted recommendations define validation and test targets.
For sequential models, we preserve item order by placing liked movies before accepted recommendations and expanding these histories into the pre-augmented RecBole sequence format~\cite{zhao2021recbole}.
The targets stay fixed while the input representation matches each model family.
Matching pools controls item availability but not model information: LLMs use raw dialogue and pretrained knowledge, whereas CF and sequential models use interaction histories.
Our comparisons therefore evaluate systems rather than reranking under identical information.

\subsection{Two-stage Pipeline and Candidate Generation}

We use a two-stage pipeline with two roles: a candidate generator returns a pool \(C_u^K\) of \(K\) catalog items for dialogue \(u\), and a reranker orders items from this pool and emits the top-10 recommendation list.
The same model can fill either role. In our experiments, CF and sequential models act as rerankers over a matched semantic pool and also produce top-\(K\) pools that LLMs then rerank.

The primary candidate generator is content-based filtering (CBF) over movie metadata.
We embed each catalog item with all-mpnet-base-v2, a Sentence-BERT model~\cite{reimers2019sentencebert}, and L2-normalize the resulting vectors \(\mathbf{e}_i\).
For dialogue \(u\) with liked set \(L_u\) and disliked set \(D_u\), we form centroids
\begin{align}
    \mathbf{c}^{+}_u = \mathrm{norm}\left(\frac{1}{|L_u|}\sum_{j\in L_u}\mathbf{e}_j\right), \qquad
    \mathbf{c}^{-}_u = \mathrm{norm}\left(\frac{1}{|D_u|}\sum_{j\in D_u}\mathbf{e}_j\right),
\end{align}
and score each catalog item \(i \notin L_u \cup D_u\) by
\begin{align}
s_{u,i} = \lambda_{+}\left\langle \mathbf{e}_i,\, \mathbf{c}^{+}_u\right\rangle \;-\; \lambda_{-}\left\langle \mathbf{e}_i,\, \mathbf{c}^{-}_u\right\rangle,
\end{align}
with \(\lambda_{+}=1\) and \(\lambda_{-}=0.5\). The negative term is dropped when \(D_u\) is empty.
The top-\(K\) items by \(s_{u,i}\) form \(C_u^K\), which we shuffle before insertion into the \texttt{<CANDIDATES>} block of LLM prompts to reduce prompt-position effects.

CBF pools of size \(K=250\) drive the main reranker comparison.
For candidate-pool sensitivity, we also evaluate zero-shot generation (no candidate list), CBF pools at \(K \in \{500, 1000\}\), and full-catalog reranking, where all 6,924 catalog titles are placed in a single prompt.
Additionally, we build top-250 pools with EASE as the CF retriever and SASRec as the sequential retriever for the same LLM rerankers, isolating first-stage effects.

\subsection{Rerankers, Inference, and Evaluation}

\paragraph{Rerankers}
We compare proprietary API LLMs, open-weight LLMs, and a fine-tuned open-weight LLM against CF and sequential rerankers.
LLM rerankers receive the dialogue context and candidate titles then generate an ordered list of movie titles.
CF rerankers score the same candidate items using collaborative signals learned from the RecBole interaction data~\cite{zhao2021recbole}.
Sequential rerankers score candidates using the corresponding ordered dialogue/user sequence.
We also include the unreranked CBF order and a popularity baseline.
The fine-tuned reranker is Qwen2.5-7B-Instruct adapted with LoRA (rank 16, \(\alpha=32\), 3 epochs, learning rate \(2\!\times\!10^{-5}\)) on the c250 training prompts.
For the main reranking runs, LLMs are decoded with temperature 0 and top-p left at its default of 1.0. The repository contains full prompt templates, exact provider model identifiers, run configurations, outputs, and parsing code.

\paragraph{Metrics}
Our primary metric is NDCG@10~\cite{jarvelin2002cumulated} with binary relevance over the accepted target movies.
We also report Hit@10, item coverage@10, and average training-set popularity@10.
LLM outputs are parsed as ordered title lists and matched to ReDial catalog items after Unicode-normalized, lowercased title matching with trailing punctuation stripped.
Titles that cannot be matched to a catalog item are kept in the raw artifact but receive no metric credit.
In candidate-constrained settings, the prompt instructs the LLM to choose only from the candidate list.
Generated titles outside that list receive no metric credit, even if they match a catalog item.

\paragraph{Retrieval diagnostics and significance}
CandRecall@250 is the mean fraction of ground-truth items retrieved into the top-250 candidate pool.
Oracle NDCG@10 is the best NDCG@10 attainable by a reranker that can only rank items from that pool.
Confidence intervals use bootstrap resampling over dialogue-level examples.
For the main reranker comparison, we test NDCG@10 against EASE with paired Wilcoxon signed-rank tests and Holm correction~\cite{holm1979simple}.

\subsection{Temperature sensitivity and list stability}
For each LLM reranker we run 20 generations per prompt at each temperature in \(\{0, 0.5, 1.0, 2.0\}\) over a fixed 30-prompt subset of the c250 test set, with top-\(p=1.0\).
Anthropic models (Claude Opus 4.6, Claude Sonnet 4.6) are capped at \(T=1.0\) by the provider.
We report mean NDCG@10 and dialogue-level variation across repeated generations.
List changes are measured with Jaccard distance@10 over the top-10 sets and position disagreement@10, the mean fraction of rank-aligned positions whose items differ across paired generations.

%% file: sections/03_results.tex
\section{Results}
\label{sec:results}

The results show three main patterns.
Proprietary LLMs lead under a fixed semantic candidate pool, first-stage retrieval can change NDCG@10 as much as model choice, and decoding temperature affects list stability more than average ranking quality.
\begin{table*}[t]
\centering
\footnotesize
\caption{Reranker effectiveness on ReDial under zero-shot scoring (c0, no candidate list) and strict candidate-aware scoring over the semantic top-250 pool (c250).
Rows group LLM, collaborative-filtering, sequential, and popularity rerankers.}
\label{tab:reranker-comparison}
\newcommand{\emptycell}{--}
\begin{tabular*}{\textwidth}{@{\extracolsep{\fill}}l cc cc cc cc@{}}
\toprule
 & \multicolumn{2}{c}{NDCG@10 $\uparrow$} & \multicolumn{2}{c}{Hit@10 $\uparrow$} & \multicolumn{2}{c}{Cov.@10 $\uparrow$} & \multicolumn{2}{c}{Pop.@10 $\downarrow$} \\
\cmidrule(lr){2-3} \cmidrule(lr){4-5} \cmidrule(lr){6-7} \cmidrule(lr){8-9}
Reranker & c0 & c250 & c0 & c250 & c0 & c250 & c0 & c250 \\
\midrule
\multicolumn{9}{@{}l}{\textsc{Semantic baseline}} \\
\quad Semantic rank~\cite{reimers2019sentencebert} & \emptycell & 0.0079 & \emptycell & 0.0260 & \emptycell & 0.4018 & \emptycell & 13.25 \\
\addlinespace[0.35em]
\multicolumn{9}{@{}l}{\textsc{Proprietary LLMs}} \\
\quad Claude Opus 4.6~\cite{anthropic2026opus46} & 0.2925$^{*}$ & 0.1497$^{*}$ & 0.4875 & 0.2713 & 0.2041 & 0.2639 & 52.12 & 42.80 \\
\quad GPT-5.2~\cite{openai2025gpt52} & 0.2427$^{*}$ & 0.1335$^{*}$ & 0.4094 & 0.2523 & 0.2163 & 0.2858 & 44.63 & 36.64 \\
\quad Claude Sonnet 4.6~\cite{anthropic2026sonnet46} & 0.2207$^{*}$ & 0.1325$^{*}$ & 0.3814 & 0.2462 & 0.2169 & 0.2754 & 45.22 & 37.79 \\
\quad GPT-4.1~\cite{openai2025gpt41} & 0.1983$^{*}$ & 0.1283$^{*}$ & 0.3453 & 0.2322 & 0.2185 & 0.3037 & 40.90 & 34.22 \\
\quad GPT-4.1 Mini~\cite{openai2025gpt41} & 0.1937$^{*}$ & 0.1021 & 0.3554 & 0.1912 & 0.1999 & 0.3033 & 52.09 & 32.69 \\
\addlinespace[0.35em]
\multicolumn{9}{@{}l}{\textsc{Open-weight LLMs}} \\
\quad Qwen2.5-7B-FT~\cite{yang2024qwen25} & \emptycell & 0.0792 & \emptycell & 0.1502 & \emptycell & 0.2711 & \emptycell & 23.22 \\
\quad Llama-3.3-70B~\cite{meta2024llama33} & 0.1154 & 0.0770 & 0.2503 & 0.1602 & 0.2114 & 0.2607 & 47.86 & 33.69 \\
\quad Gemma-2-9B~\cite{gemmateam2024gemma2} & 0.1356$^{*}$ & 0.0650 & 0.2913 & 0.1281 & 0.1932 & 0.3068 & 47.87 & 26.43 \\
\quad Llama-3.1-8B~\cite{grattafiori2024llama} & 0.0815 & 0.0503 & 0.1992 & 0.1011 & 0.1749 & 0.2402 & 40.68 & 27.03 \\
\quad Qwen2.5-7B~\cite{yang2024qwen25} & 0.0622 & 0.0500 & 0.1952 & 0.1071 & 0.2064 & 0.2620 & 57.75 & 29.62 \\
\quad Llama-3.2-3B~\cite{meta2024llama32} & 0.0603 & 0.0278 & 0.1622 & 0.0571 & 0.1989 & 0.2939 & 40.72 & 20.93 \\
\addlinespace[0.35em]
\multicolumn{9}{@{}l}{\textsc{Collaborative filtering}} \\
\quad EASE~\cite{steck2019ease} & \emptycell & 0.0939 & \emptycell & 0.2072 & \emptycell & 0.2370 & \emptycell & 59.99 \\
\quad ItemKNN~\cite{sarwar2001item} & \emptycell & 0.0876 & \emptycell & 0.1982 & \emptycell & 0.3258 & \emptycell & 43.86 \\
\quad LightGCN~\cite{he2020lightgcn} & \emptycell & 0.0833 & \emptycell & 0.2042 & \emptycell & 0.2139 & \emptycell & 59.96 \\
\quad BPR~\cite{rendle2009bpr} & \emptycell & 0.0826 & \emptycell & 0.1892 & \emptycell & 0.2096 & \emptycell & 63.60 \\
\addlinespace[0.35em]
\multicolumn{9}{@{}l}{\textsc{Sequential models}} \\
\quad SASRec~\cite{kang2018sasrec} & \emptycell & 0.0705 & \emptycell & 0.1802 & \emptycell & 0.1592 & \emptycell & 67.96 \\
\quad GRU4Rec~\cite{hidasi2016gru4rec} & \emptycell & 0.0703 & \emptycell & 0.1792 & \emptycell & 0.2301 & \emptycell & 57.82 \\
\quad NARM~\cite{li2017narm} & \emptycell & 0.0686 & \emptycell & 0.1612 & \emptycell & 0.2402 & \emptycell & 58.82 \\
\quad SRGNN~\cite{wu2019srgnn} & \emptycell & 0.0615 & \emptycell & 0.1592 & \emptycell & 0.1853 & \emptycell & 64.44 \\
\addlinespace[0.35em]
\multicolumn{9}{@{}l}{\textsc{Popularity baseline}} \\
\quad Pop~\cite{zhao2021recbole} & \emptycell & 0.0388 & \emptycell & 0.1061 & \emptycell & 0.0477 & \emptycell & 107.68 \\
\bottomrule
\end{tabular*}
\vspace{0.25em}
\begin{minipage}{\textwidth}
\footnotesize
Cov.~is item coverage of the top-10 list.
Pop.~is mean training-set popularity of top-10 items.
$^{*}$ marks NDCG@10 significantly above the EASE c250 baseline by paired Wilcoxon signed-rank test with Holm correction ($p<0.05$).
c0 and c250 LLM scores are tested separately against that baseline.
CF, sequential, and popularity baselines require a candidate set and have no c0 entry.
Qwen2.5-7B-FT was fine-tuned on c250 prompts and is reported only for that policy.
\end{minipage}
\end{table*}

\paragraph{RQ1.}
Under the matched semantic top-250 pool (Table~\ref{tab:reranker-comparison}), Claude Opus 4.6 achieves the highest strict NDCG@10 at 0.1497, with GPT-5.2, Claude Sonnet 4.6, and GPT-4.1 clustered behind between 0.1283 and 0.1335.
These four proprietary models are the only rerankers significantly above EASE (0.0939) after Holm correction.
GPT-4.1 Mini is numerically higher than EASE but not significant, and every open-weight reranker falls below EASE.
Zero-shot scoring changes this picture.
Claude Opus 4.6 reaches 0.2925, nearly twice its strict c250 score, and additional models become significant against the EASE baseline.
Table~\ref{tab:reranker-comparison} also shows differences in coverage and popularity.
Under c250, GPT-4.1 and GPT-4.1 Mini have the highest coverage among proprietary LLMs, while EASE has lower coverage and higher average popularity.
The strongest proprietary rerankers improve NDCG without relying only on the most popular items.
Open-weight models show mixed behavior: some cover a broad part of the catalog but still rank less accurately.

\begin{figure}[t]
\centering
\includegraphics[width=\columnwidth]{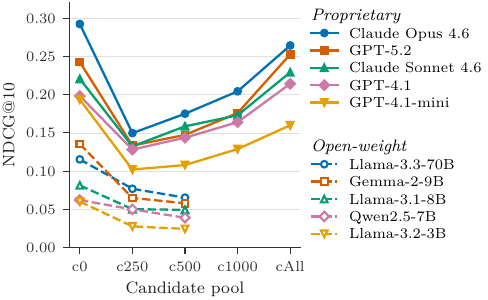}
\caption{NDCG@10 as a function of candidate-pool size across LLM rerankers with semantic retrieval.
Settings span c0 (zero-shot, no candidates), strict candidate-aware top-$k$ scoring at c250, c500, and c1000, and full-catalog (cAll).}
\Description{Line plot of NDCG at 10 across c0, c250, c500, c1000, and cAll candidate settings for the evaluated LLMs.}
\label{fig:candidate-pool-sensitivity}
\end{figure}

\paragraph{RQ2.}
Figure~\ref{fig:candidate-pool-sensitivity} shows how candidate access reshapes the same rerankers.
Every proprietary model gains substantially as the pool expands.
Strict NDCG@10 rises by 57\% (GPT-4.1 Mini) to 89\% (GPT-5.2) between c250 and cAll, and cAll is the best strict setting for each.
c0, scored without a candidate list, lands in the same range and slightly exceeds cAll for Opus (0.2925 vs 0.2644) and GPT-4.1 Mini (0.1937 vs 0.1600).
Open-weight rerankers do not benefit similarly, with NDCG dropping from c250 to c500 for every evaluated open-weight model.
Because pool size simultaneously changes target availability, candidate composition, prompt length, and ranking difficulty, this experiment measures their combined pipeline effect rather than isolating reranking ability.

\begin{table}[t]
\centering
\footnotesize
\caption{Effect of the first-stage retriever on LLM reranking with a fixed top-250 candidate pool.}
\label{tab:retrieval-strategy}
\begin{tabular*}{\columnwidth}{@{\extracolsep{\fill}}l c ccc@{}}
\toprule
 & & \multicolumn{3}{c}{NDCG@10 $\uparrow$} \\
\cmidrule(lr){3-5}
Retriever & CandR@250 $\uparrow$ & Oracle & Claude Opus 4.6 & Llama-3.3-70B \\
\midrule
Semantic & 0.2877 & 0.3165 & 0.1497 & 0.0770 \\
EASE & 0.4862 & 0.5222 & 0.2277 & 0.1223 \\
SASRec & 0.5093 & 0.5423 & 0.2119 & 0.1180 \\
\bottomrule
\end{tabular*}
\vspace{0.25em}
\begin{minipage}{\columnwidth}
\footnotesize
CandR@250 is the mean fraction of ground-truth movies retrieved into the candidate pool.
Oracle NDCG@10 is the best NDCG@10 achievable by a perfect reranker restricted to that pool, and upper-bounds the downstream reranker columns.
\end{minipage}
\end{table}

\paragraph{RQ3.}
Table~\ref{tab:retrieval-strategy} changes the first-stage retriever while holding candidate-pool size and reranker fixed.
Semantic retrieval places only 28.8\% of relevant items in the top-250 pool, versus 48.6\% for EASE and 50.9\% for SASRec.
Oracle NDCG@10 follows the same order.
Retriever choice carries through to the reranker.
Switching from semantic to EASE lifts Claude Opus 4.6 by 52\% (0.1497 to 0.2277) and Llama-3.3-70B by 59\% (0.0770 to 0.1223).
SASRec has the highest recall and oracle ceiling, while EASE yields the best NDCG for both rerankers.
Retrieval opportunity alone therefore does not determine downstream performance, though this design does not separate candidate composition from reranker--pool compatibility.

\begin{figure}[t]
\centering
\includegraphics[width=\columnwidth]{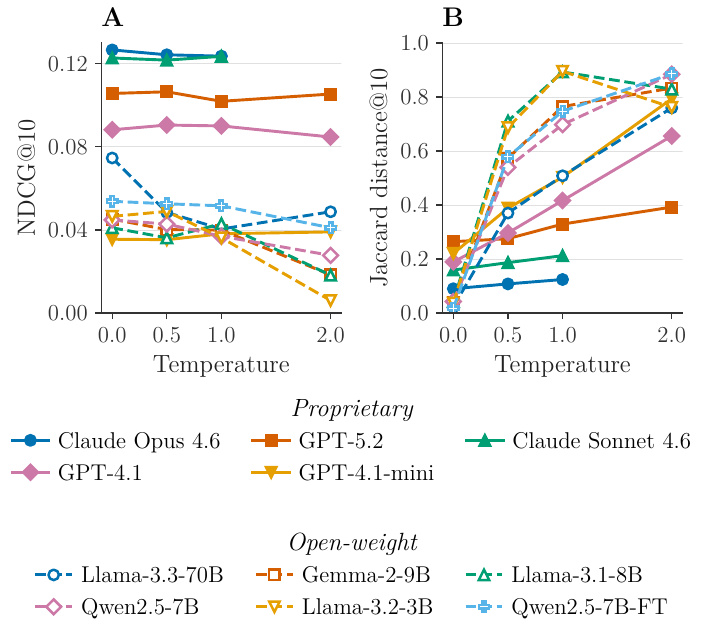}
\caption{NDCG@10 (A) and Jaccard distance@10 (B) as functions of decoding temperature on the fixed semantic c250 stability subset of 30 prompts.}
\Description{Two-panel line plot showing strict NDCG at 10 and Jaccard distance at 10 as a function of decoding temperature for all evaluated LLM rerankers.}
\label{fig:temperature-stability}
\end{figure}

\paragraph{RQ4.}
Figure~\ref{fig:temperature-stability} reports the stability subset of 30 prompts, so absolute NDCG levels are not comparable to the full-test numbers in Table~\ref{tab:reranker-comparison}.
Across API models, mean strict NDCG@10 is roughly flat in temperature while Jaccard distance@10 rises with each step.
The Anthropic models cap at $T=1.0$ by the provider, so their traces stop there.
Claude Opus 4.6 is the most stable.
Its Jaccard distance moves from 0.0900 at $T=0$ to 0.1240 at $T=1.0$, position disagreement@10 rises from 0.2700 to 0.3500, and NDCG@10 changes little.
Open-weight rerankers are less stable.
Llama-3.3-70B begins almost deterministic at $T=0$ (Jaccard 0.0230, position disagreement 0.0250) and reaches 0.7600 and 0.7930 at $T=2.0$.
Its NDCG@10 also falls from 0.0750 to 0.0490.

%% file: sections/04_discussion.tex
\section{Discussion and Conclusion}
\label{sec:discussion}

\paragraph{Retrieval choices are as important as the reranker.}
Switching first-stage retrievers improves NDCG@10 by 52--59\% across both rerankers, while larger candidate pools also substantially improve proprietary models. Yet the higher-recall SASRec pool trails EASE after reranking, showing that candidate availability, composition, and reranker--pool compatibility jointly determine downstream quality. Treating candidate generation as an implementation detail thus conflates pipeline and reranker quality.

\paragraph{Candidate access and scoring policy shape headline LLM gains.}
Claude Opus 4.6 reaches 0.2925 in zero-shot generation but 0.1497 under strict candidate-aware scoring on the matched semantic top-250 pool.
Because c0 and c250 differ in both candidate access and scoring policy, this contrast does not isolate either factor.
It instead shows how strongly the evaluation protocol affects the apparent advantage over EASE.
Under the matched c250 pool, only four proprietary models significantly outperform EASE, while every open-weight reranker falls below it.
These are system-level comparisons rather than tests of reranking under identical information.

\paragraph{Stability is a separate axis from accuracy.}
Mean accuracy and list identity respond differently to decoding temperature.
For the strongest proprietary models, NDCG@10 changes little between temperature 0 and 1, while Jaccard distance@10 and position disagreement@10 increase.
The model can therefore return lists with similar accuracy but different items and rankings.
Llama-3.3-70B is more temperature-sensitive, with Jaccard distance@10 rising from 0.0230 to 0.7600 and NDCG@10 falling from 0.0750 to 0.0490.
Reporting only mean NDCG hides this deployment-relevant behavior.

\paragraph{Scope and future work.}
Our experiments use ReDial, a single movie-domain CRS benchmark with a 6,924-item catalog.
Future work will test whether these findings generalize to larger catalogs, other domains, and conversational settings.
We did not systematically measure cost or latency.
Future evaluations should compare quality, cost, and latency, particularly for large and full-catalog candidate pools.

\paragraph{Toward a reporting norm.}
LLM-based CRS evaluations should report the retriever, candidate-pool size, scoring policy for off-candidate and unmatched generations, decoding configuration, prompt template, and exact model and provider identifier.
Headline results should be interpreted together with this protocol because pipeline choices can affect measured gains as much as model choice.